\documentclass{article}
\usepackage{PRIMEarxiv}

\pdfoutput=1 

\usepackage{amsmath} 
\usepackage{amssymb}  
\usepackage[utf8]{inputenc} 
\usepackage[T1]{fontenc}    
\usepackage{hyperref}       
\usepackage{url}            
\usepackage{booktabs}       
\usepackage{amsfonts}       
\usepackage{nicefrac}       
\usepackage{microtype}      
\usepackage{fancyhdr}       
\usepackage{graphicx}       
\graphicspath{{media/}}     

\title{An innovative Deep Learning Based Approach for Accurate
Agricultural Crop Price Prediction
}

\author{
  Mayank Ratan Bhardwaj and Jaydeep Pawar \\
  Indian Institute of Science \\
  Bengaluru\\
  \texttt{\{mayankb, jaydeeppawar\}@iisc.ac.in} \\
   \And
  Abhijnya Bhat \\
  PES University \\
  Bengaluru\\
  \texttt{abhijnya.bhat@gmail.com} \\
  \And
  Deepanshu, Inavamsi Enaganti, 
  Kartik Sagar and Y. Narahari \\
  Indian Institute of Science \\
  Bengaluru\\
  \texttt{\{deepanshu1, inavamsie, kartiksagar, narahari\}@iisc.ac.in}
}

\begin{document}
\maketitle

\begin{abstract}
Accurate prediction of agricultural crop prices is a crucial input for decision-making by various stakeholders in agriculture: farmers, consumers, retailers, wholesalers, and the Government. These decisions have significant implications including, most importantly, the economic well-being of the farmers. In this paper, our objective is to accurately predict crop prices using historical price information, climate conditions, soil type, location, and other key determinants of crop prices. This is a technically challenging problem, which has been attempted before. In this paper, we propose an innovative deep learning based approach to achieve increased accuracy in price prediction. The proposed approach uses graph neural networks (GNNs) in conjunction with a standard convolutional neural network (CNN) model to exploit geospatial dependencies in prices. Our approach works well with noisy legacy data and produces a performance that is at least 20\% better than the results available in the literature. We are able to predict prices up to 30 days ahead. We choose two vegetables, potato (stable price behavior) and tomato (volatile price behavior) and work with noisy public data available from Indian agricultural markets.
\end{abstract}


\section{Introduction}
Accurate predictions of crop yield and crop price provide valuable inputs for decision making by various stakeholders in agriculture: farmers, consumers, retailers, wholesalers, dealers, and the Government. Some of these decisions have far-reaching implications for the  economic well-being of farmers, ensuring food security, stability of supplies, breeding of seeds, and for economic planning in general. This paper focuses on the problem of crop price prediction. 

There are several factors affecting crop prices. These include the expected yield, expected demand,  export projections, import decisions, supply chain factors, weather conditions, geospatial factors,  unanticipated events such as a pandemic or a flood, etc. Compounding this is the fact that the data that are available in many emerging economies about historical crop prices and crop price variations have several issues such as missing values, outliers, and even data entry errors. Accurate prediction of crop prices is therefore a grand challenge problem but at the same time an important one to help secure the economic prosperity of farmers. This paper focuses on how geospatial dependencies can be harnessed to obtain improved accuracy in  predictions of crop prices.
\subsection{Use-Cases for Price Prediction}
Small and marginal farmers face many challenges in understanding and interpreting the market and its price dynamics to use them to their advantage. With price forecasts and simple advisories available, the farmers will be able to decide when to harvest, when to  sell their crops, how to plan for the next season, etc.  
There are many potential benefits of knowing what the prices are likely to be and we describe two use-cases below.

\subsubsection{A Use-Case for the Farmer: When to Harvest} 
Given any crop, the correct time of harvest is crucial in preventing crop losses. 
Losses can be caused by field animals, plant diseases, insect pests, or certain weather conditions. 
There is a small window of time during which harvesting produces the best results. This window is different for different crops. Closer to the harvesting time window, there could be uncertainties such as changing weather conditions and other external factors.  {\em When to harvest\/} is therefore a crucial decision that the farmer is faced with and the market price is a key determinant for this decision. If accurate price predictions are available for the next several days and weeks, the farmer will be able to make an informed decision. Otherwise, the farmer may end up incurring heavy labor costs,  cold storage costs, and transportation costs.

\subsubsection{A Use-Case for the Government: Interventions to Support Farmers}
If the one-week or two-week price prediction for a certain crop forecasts attractive prices, then the farmers are elated. However, if low prices are predicted, the farmers could get into panic mode. In such situations, the Government can intervene and assure the farmers that it would buy the produce (or at least a part of it) at a price that is favorable to the farmer. The Government can also initiate various measures for supporting the farmers such as connecting the farmers to food processing units, enabling affordable storage of the produce, regulating import and export, etc.  


\subsection{Contributions and Outline}
Agricultural crop price prediction is a technically challenging problem with an extensive body of literature. This paper advances the state-of-the-art by improving the accuracy of predictions further by introducing innovations in modeling. It uses geospatial proximity in addition to temporal data for making better price predictions. In particular, we use graph neural networks (GNNs) in conjunction with standard deep learning models such as convolutional neural networks (CNNs) to exploit any geospatial dependencies in crop prices.  
GNNs have powerful representation learning capabilities for graph structured data and have been applied in wide-ranging applications \cite{ZHOU20}. 
In the case of agricultural crop prices, it is observed that neighbouring mandis experience similar weather patterns, farming practices, and soil conditions due to physical proximity. In addition to these factors, produce movement between nearby mandis causes a semblance of price stabilisation in mandis within a certain geographical distance of each other. 
We represent markets as the nodes of the graph and we use the edges to capture their proximity.
The data that we use pertains to all the markets in India and covers two crops from the opposite ends of the spectrum of price volatility, tomato and potato.


Our first experiments are with tomato. 
We compare the results obtained for tomato with two approaches - PECAD \cite{GUO20} and CGNN (our approach). The paper \cite{GUO20} only reports coefficient of variation results for 4 days, 6 days, and 9 days. It does not report CoV for other time horizons or any other performance metric for any time horizon.    The results  clearly show that the CGNN approach outperforms the PECAD approach in all the cases reported in \cite{GUO20}. 

We then turned our attention to potato and obtained results with two approaches - PECAD \cite{GUO20} and CGNN (our approach). The paper \cite{GUO20} does not report any results for potato, so we ran the PECAD code (available from the authors of \cite{GUO20}) on our potato dataset. We computed five different performance measures: root mean square error, mean absolute error, coefficient of variation, R2 value, and Pearson's correlation coefficient. The results clearly show that the CGNN approach outperforms the PECAD approach on all performance metrics for all time horizons. 

In Section \ref{sec:Methodology}, we describe the details of our methodology: nature of data used by us; data imputation and data curation; deep learning models used; and the architecture of the price prediction network. In Section \ref{sec:Results}, we report the results from our experiments on tomato and potato crops. 

Tomato and Potato are two representative crops we have used. The same methodology is applicable to other crops as well.  

\section{Review of Relevant Work}
There have been numerous research efforts toward crop price prediction. The problem offers a number of technical challenges since the crop price is determined by a large number of factors. Due to space constraints, we only review relevant, recent efforts. First, we wish to mention that  the price prediction problem  is related to but also different from the crop yield prediction problem.  There is abundant literature on the crop yield prediction problem as well and we point the reader to a comprehensive survey on yield prediction \cite{KLOMPENBURG20}. Here, we review recent papers on price prediction and a paper on yield prediction that are most relevant to our work.

Ma, Nowocin, Marathe, and Chen \cite{MA19} present a crop price forecasting system using data from Agmarknet (a Government website) in India.
They train a classification model of prices using 1352 markets in India and produce interpretable price forecasts. 
The pricing data available from Agmarknet is sparse and the authors impute missing entries using collaborative filtering to obtain a dense dataset. Using these data, a decision-tree-based classifier is trained to  predict the direction of movement in crop prices at different markets. The system uses adaptive nearest neighbor methods to obtain
interpretable forecasts. 
The forecasting system is used to predict price changes for six crops (brinjal, cauliflower, pointed gourd, mango, tomato, and green chilli). The authors do not take into account  spatio-temporal dependencies of crop prices. Additionally, they classify the price as either going up or down, while we predict the actual prices, making comparison with their paper infeasible. 

The paper by Guo, Woodruff, and Yadav \cite{GUO20} presents a deep learning based algorithm PECAD (Price Estimation for Crops using the Application of Deep Learning), for accurate prediction of future crop prices based on past pricing and volume patterns. The paper uses real-world daily price and volume data of different crops across India
using the same database as above, Agmarknet. From the available markets, they eliminate the markets that have less than 10\% data available and carry out experimentation on the remaining markets.
The data are batched into $n$-day nonoverlapping snippets and pre-processed using imputation techniques such as SoftImpute to account for missing data entries.  PECAD works with a wide and deep neural network architecture consisting of two separate convolutional neural network models (trained for pricing and volume data respectively). PECAD outperforms existing  baseline methods by achieving  25\% less Coefficient of Variation (CoV). This model takes into account the latitude and longitude of the large number of markets to capture the effect of geospatial factors at a high level. It does not specifically exploit the dependence of crop prices at one market on the prices at other markets.

The paper by Madaan et al. \cite{MADAAN19} looks at onion and potato trading in India and presents an evaluation of a price forecasting model, and an anomaly detection and classification system to identify incidents of hoarding of stock by the traders. The dataset consists of a time series of wholesale prices and arrival volumes of the agricultural commodities at several village-level marketplaces, and retail prices of the commodities at the city centers. The paper presents 
a qualitative analysis of the effect on these time series of events such as hoarding, weather disturbances, and external shocks. The authors employ ARIMA and SARIMA models in conjunction with an LSTM model for their prediction tasks. The pricing models are useful to reduce information asymmetries and to detect  anomalies that can help regulate agricultural markets to operate more fairly. This paper does not take into account factors such as geospatial dependencies.

The authors  of \cite{ZHANG20} propose a  model selection framework that  includes time series features and forecast horizons for predicting the price of agricultural commodities. As many as 29 features are considered as possible influencers of prices.  Three models are explored: neural networks, support vector regression, and extreme learning machine. Random forests and support vector machines are applied for learning the relationships between the features. Techniques are proposed to remove feature redundancies. This model does not capture any geospatial dependencies. 

In \cite{JAIN22}, the authors present the architecture of an end-to-end pipeline for robust crop price prediction through the analysis of  historical marketplace data and weather data. They also discuss data quality-related features. The paper presents a framework that facilitates  context-based model selection strategies with data quality, model stability, historical price trends as the context determinants. The authors  experiment with various regression models and show the results for tomato and maize crops for 14 markets in the state of Karnataka, India. This modeling effort  also does not capture geospatial factors. 

The  paper by Fan et al.  \cite{FAN22} explores the use of graph neural networks (GNN) in conjunction with  recurrent neural networks (RNN) to capture  geospatial as well as temporal knowledge into crop yield prediction. The work is able to take advantage of the spatial structure in the data to produce better yield predictions.  Their model first extracts year-wise embeddings using a CNN model. These are then fed as input to the GNN, which enables inductive representation learning on graphs. These outputs are the fed into the RNN to obtain the final yield results. The graph here is an unweighted graph that consists of more than 2000 counties as nodes and the edges represent neighborhood relationships among the counties. Detailed experiments on large-scale datasets covering 41 states in the United States demonstrate that the approach outperforms state-of-the-art machine learning methods across multiple datasets by almost 10 \%. The graph captures the neighborhood relationship among the counties, thereby bolstering the results. It should be noted here that \cite{FAN22} works on yield prediction in USA, which has rich and perfect data. Our work, on the other hand, deals with crop price prediction in India, where the data available are sparse as well as noisy.

As already pointed out, the crop yield prediction is a different problem than the crop price prediction problem. The set of factors affecting crop price form a superset of factors affecting crop yield. Crop price crucially depends on demand and supply (which in turn depends on yield) while yield depends very little  on price or demand.
Price prediction is clearly a more involved problem and taking into account geospatial dependencies is in itself an interesting, independent problem. 

\section{METHODOLOGY} \label{sec:Methodology}
In this section, we describe the methodology used to predict the daily price 
for agricultural markets.  These markets are popularly called {\em mandis\/} in India and we use that nomenclature in the rest of the paper.  It should be noted that a mandi  can be characterized
 by its latitude and longitude.  We start by describing the data used and pre-processing techniques applied on the data. 

\subsection{Crop and Weather Data}\label{subsec:Data}
The website run by the Directorate of Marketing \& Inspection (DMI), Ministry of Agriculture and Farmers Welfare, Government of India\footnote{\url{http://agmarknet.gov.in/}} provides daily Price (in Rs. per Quintal) and Arrival Data (in Tonnes) for all mandis in India (1 quintal = 100 kilograms and 1 Tonne = 1 megagram). The Price data includes the daily maximum price, minimum price, and modal price for each crop in each mandi. However, data availability is highly variable across crops and mandis. 
For our experimentation, mandis with very sparse data were weeded out and from the 1320 and 730 mandis with tomato and potato data, the top 557 and 676 mandis, respectively, were considered. Only those mandis were considered that had price and arrival data for at least 4 days in each year between 2014 and 2018.

Weather data consisting of hourly values of Rainfall, Temperature, Surface net solar radiation, and Humidity were taken from Copernicus, the European Union's Earth Observation Programme\footnote{\url{https://climate.copernicus.eu}} for a period of five years (2014-2018). This data could be used directly as all the missing values and anomalies were corrected before the data was uploaded.

\subsection{Pre-Processing}\label{subsec:Pre-Processing}
\begin{figure}[h]
        \centering
        \framebox{\parbox{6in}{\includegraphics[scale=0.56]{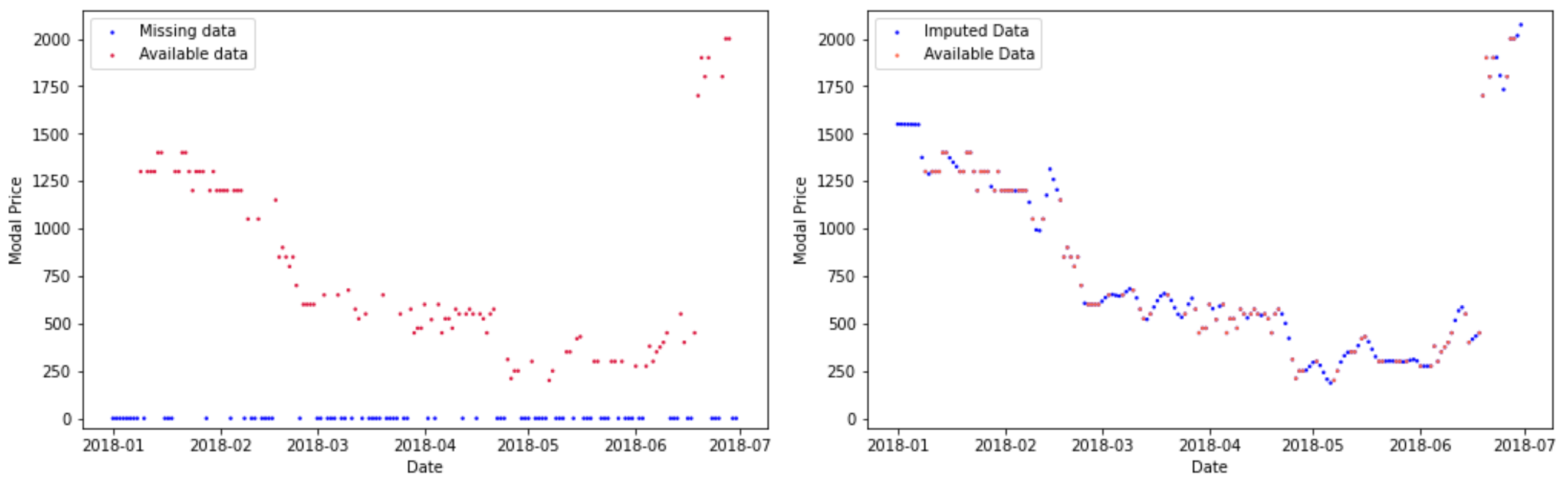}}}
        \caption{Tomato prices, before imputation on the left, and after imputation on the right, for the first half of 2018 in Sardhana mandi. The zero-prices correspond to the dates where datum was not available.}
        \label{fig:imputedTomatoPriceSardhana}
    \end{figure}

The price and arrival data have many missing values. An illustrative example of sparsity in tomato price data in the Sardhana mandi of Meerut district is provided in Fig. \ref{fig:imputedTomatoPriceSardhana} (left side). All over India, 46\% of the price data are missing. While some mandis like Kharar have 1.81\% data missing, mandis like K.R.Pet have 98.14\% data missing.

In addition to the missing values, we found several instances where the data was misreported, probably due to the manual data-entry process. In these cases, the entry for a particular day would typically have an additional zero or a missing digit. To catch these and other outliers, we checked if the current value was either six times higher or lower than a sixth of the average of the previous week. If there were no values present in the previous week, we backtracked until the most recent ground-truth value available.

We then performed year-wise spline imputation \cite{AHLBERG16} on these data. Since spline does not work well for sparse data, this technique was used only for mandis that had data available for more than half of the year. 
We found that cubic splines work best for these data.

The spline imputation also gives rise to some outliers, which were caught using a moving-window approach. For each imputed data-point, a window of size 14 (one week before and one week after) was taken and all the ground truth values in that window were considered. If no ground-truth values were present in that window, it was expanded on both sides until sufficient values were obtained. Then, the imputed value was reported as an outlier if it was beyond +/-15\% of the window's extrema.

Finally, we used linear interpolation to impute the remaining missing values.
Fig. \ref{fig:imputedTomatoPriceSardhana} displays the tomato prices in Sardhana mandi, from January 2018 to July 2018, before imputation (on the left) and after imputation (on the right).

All numerical values were normalised. The date feature was broken down into two cyclic features, namely month and day, both of which were transformed into two dimensions using a sine and cosine transformation. This is done using the following (circular embedding) transformation for the features month as well as day of month : $x_{sin} = \sin{(2 \pi x/\max(x))}, x_{cos} = \cos{(2 \pi x/\max(x))}$.
Since all the mandis are closed on Sundays, resulting in no activity, all Sundays were removed from the data set. The latitude and longitude of each mandi were also provided as additional features.

Price data are susceptible to minor fluctuations even over a short span of a few days. These fluctuations occur due to a variety of factors including measuring errors, human behavior, and randomised natural phenomena, which cannot be measured due to lack of data and set mechanisms. To smoothen the data, the features were averaged over the duration of a few days. The corresponding data entry obtained by averaging all the features over $n$ days is called an \textit{$n$-day snippet}. A single $n$-day snippet is categorised
by crop, mandi, and the first day of the snippet. Thus the data corresponding to a crop in a particular mandi was broken into continuous intervals of $n$-day snippets after the removal of Sundays. All $16$ features for each $n$-day snippet were taken as input. $4$, $6$, $9$, $15$, and $30$ were considered as possible values for $n$.  A list of all the features is provided in Table \ref{table:Parameters}.

\begin{table} [h]
\begin{center}
\begin{tabular}{|c c c c|}
\hline
\textbf{Features from}  & \textbf{Features from}  & \textbf{Temporal \&} & 
\textbf{Computed} \\ 
\textbf{Agmarknet}  & \textbf{Copernicus}  & \textbf{Spatial Features} & 
\textbf{Features} \\ 
\hline
Modal Price & Temperature & Day & Day of month - Cyclic encoding: sine\\ 
Arrival Quantity &  Total Precipitation & Month  & Day of month - Cyclic encoding: cosine \\ 
 & Surface net Solar Radiation &  Year  & Month - Cyclic encoding: sine \\ 
 & Relative Humidity  &  Latitude & Month - Cyclic encoding: cosine \\ 
 & & Longitude & Previous 7 days average of modal price\\ \hline
\end{tabular}
\end{center}
\caption{Input features}
\label{table:Parameters}
\end{table}


Model training was carried out over the 3-year period of 2014-2016. Validation was done using the data for year 2017 and the data for year 2018 was used for testing.

\subsection{Deep Learning Models}

We now present the training process for our CGNN model. $n$-day snippets were used to predict the average price of the subsequent $n$-day snippet. 

The weather features are available as hourly values for each mandi. A 1-dimensional convolution (1D CNN) operation was applied to capture temporal changes and generate a compact embedding. A 1D CNN is very effective when one expects to derive embeddings from shorter, fixed-length sequences. The 1D-CNN was paired with a ReLU activation function and a Max pool layer. Four such instantiations were created; one for each weather feature. The outputs of the 4 models were then concatenated with the remaining data. As the number of intermediate features obtained at this stage was large, the concatenated result was passed through two successive Fully-Connected layers. The resulting embedding vector of size 5 for each mandi constitutes the input to the GNN model.

\begin{figure}
        \centering
        \framebox{\parbox{3in}{\includegraphics[scale=0.45]{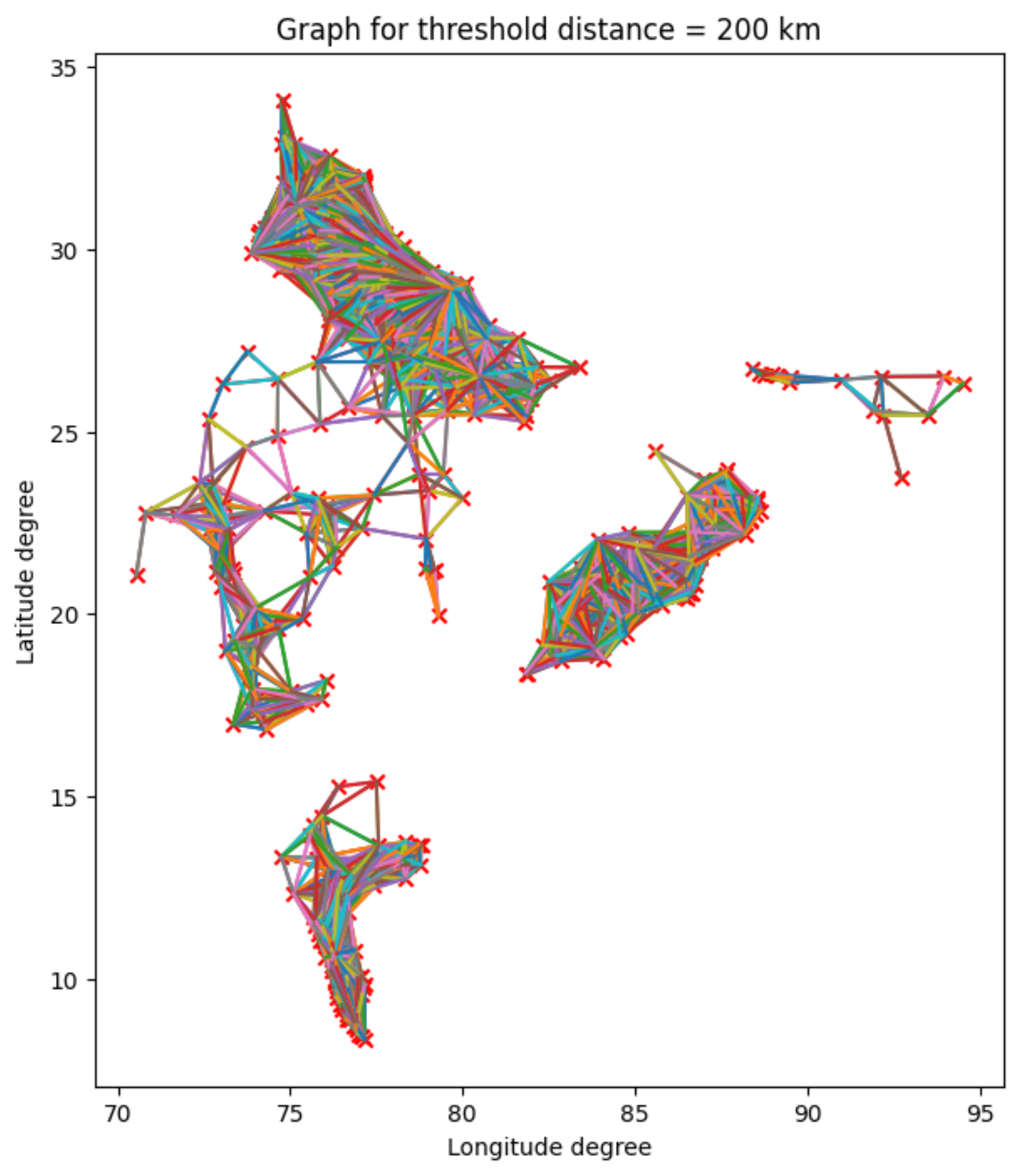}}}
        \caption{Graph of mandis where tomato is sold. An edge exists between two mandis if they are within 200 Km of each other.}
        \label{fig:tomatoGraph}
    \end{figure}

For the GNN model, the graph network was built for all the mandis in the data set that were shortlisted (on the basis of data availability) for the crop being considered. Each vertex of the graph represents a shortlisted mandi and is categorised by the mandi’s geographic location given by its latitude and longitude. If the direct distance (as the crow flies) between two mandis was found to be within a certain threshold, an edge was added to the two vertices in the graph corresponding to these mandis. Various such thresholds were considered. Of these, 200 Km provided the best prediction results. It should be noted that each crop would have a different subset of shortlisted mandis (depending on data availability) and thus, a separate graph would be required for each crop. Furthermore, the perishability, growth requirements, storage and price dynamics of each crop could result in a different threshold being selected for each crop. Fig. \ref{fig:tomatoGraph} shows the final graph obtained for the threshold of 200 Km for tomato. 


\begin{figure}[thpb]
        \centering
        \framebox{\parbox{6in}{\includegraphics[scale=0.2]{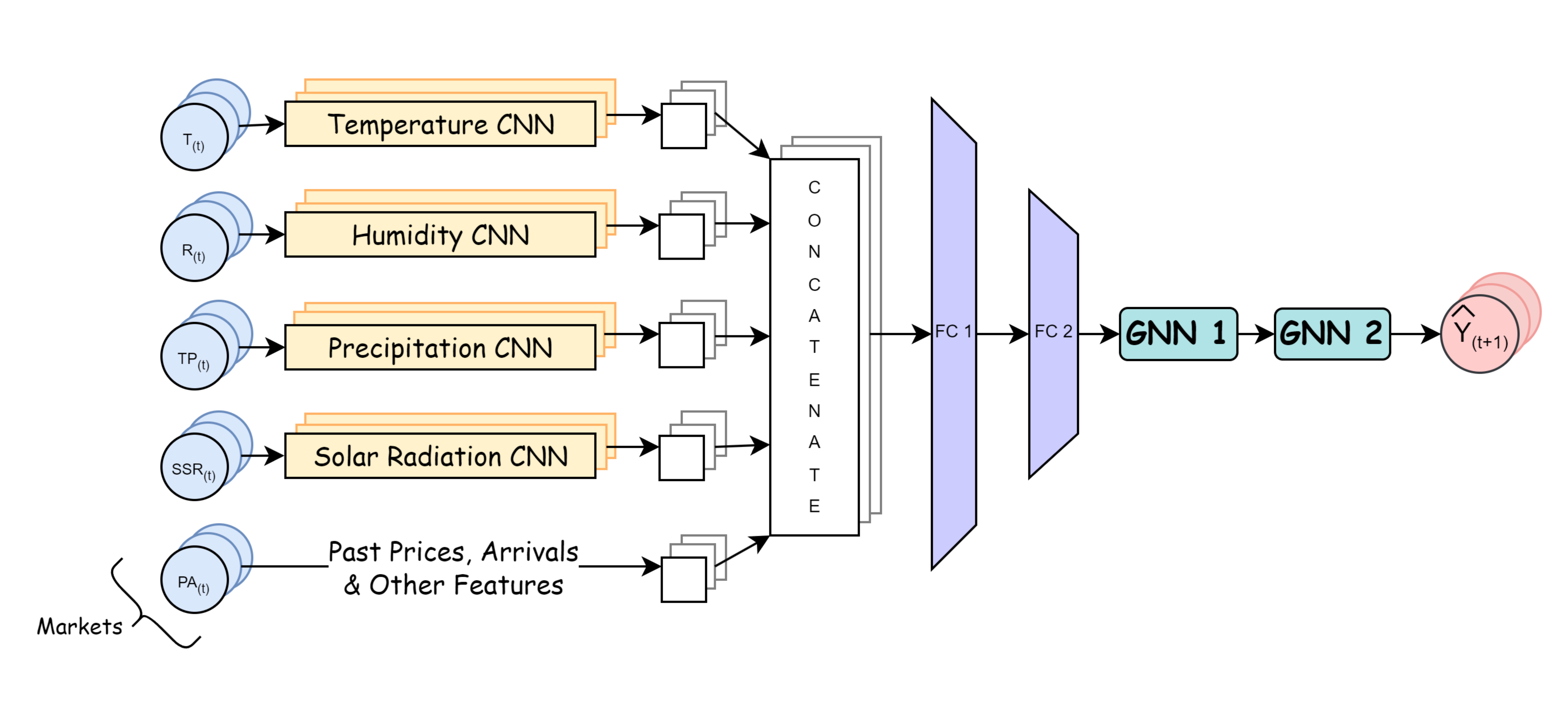}}}
        \caption{CGNN Model}
        \label{fig:model}
    \end{figure}

The GNN model consisted of two GraphSAGE \cite{HAMILTON17} layers followed by ReLU activation layers. The detailed architecture of the model is shown in Fig. \ref{fig:model}.

\subsection{Performance Metrics}
Since two distributions cannot be compared using a single parameter, it is impossible to measure the goodness of fit of predicted values using a single metric. Similarity or prediction accuracy is measured in different dimensions by different metrics. In the absence of a set of metrics that can be used for a universal comparison, it is best to use as many metrics as possible. We found that all papers on crop price prediction in the literature were reporting a single or at most two metrics. In order to confirm the veracity of our models, we have computed and reported all of the metrics mentioned below.  
\begin{itemize}
    \item Mean Absolute Error (\textbf{MAE}) is calculated as the sum of absolute errors between predicted price and target price, divided by the sample size.
    \begin{equation}
        MAE = \frac{\Sigma_{i=1}^n \left| y_i - x_i \right|}{n}
    \end{equation}
    \item Root Mean Square Error (\textbf{RMSE}) is the standard deviation of the prediction errors.
    \begin{equation}
        RMSE = \sqrt{\frac{\Sigma_{i=1}^n (y_i - x_i)^2}{n}}
    \end{equation}
    \item Coefficient of Variation (\textbf{CoV}) is calculated as the RMSE divided by the mean of the sample size.
    \begin{equation}
        CoV = \frac{RMSE}{\Bar{x}} = \frac{\sqrt{\Sigma_{i=1}^n (y_i - x_i)^2 /n}}{\Bar{x}}
    \end{equation}
    \item R2 score (\textbf{$R^2$}) is defined as the proportion of the variance in the dependent variable that is predictable from the independent variables. 
    \begin{equation}
        R^2 = 1 - \frac{\Sigma_{i=1}^n (y_i - x_i)^2}{\Sigma_{i=1}^n (y_i - \Bar{y})^2}
    \end{equation}
    \item Pearson's Coefficient ($\textbf{r}$) is a measure of linear correlations between the predicted prices and target prices. It is calculated as the covariance between the two sets divided by the product of their standard deviations.
    \begin{equation}
        r = \frac{\Sigma_{i=1}^n (x_i - \Bar{x}) (y_i - \Bar{y})}{\sqrt{\Sigma_{i=1}^n (x_i - \Bar{x})^2} \sqrt{\Sigma_{i=1}^n (y_i - \Bar{y})^2}}
    \end{equation}
\end{itemize}

\section{EXPERIMENTAL RESULTS FOR TOMATO AND POTATO CROPS} \label{sec:Results}
In this section, we present the results of our experimentation on two popular crops in India: tomato and potato.  We chose India because of its predominantly agrarian character with more than half of the workforce involved in agriculture.
We chose tomato and potato for a clear reason: prices of potato are somewhat stable while the prices of tomato are volatile.  

The CGNN model was trained for 4000 epochs and the best checkpoints were saved based on the performance of the model on the validation set. A learning rate of 0.1 was used, coupled with a scheduler that reduced the learning rate to 0.01 after 3000 epochs.

\subsection{Results for Tomato}
Table \ref{table:TomatoAY} shows the results obtained for tomato with two approaches - PECAD \cite{GUO20} and CGNN (our approach). The paper \cite{GUO20} only reports coefficient of variation results for 4 days, 6 days, and 9 days. It does not report CoV for other time horizons or any other performance metric for any time horizon.  We have computed all five different performance measures.  The results in Table \ref{table:TomatoAY} clearly show that the CGNN approach outperforms the PECAD approach in all the cases reported in \cite{GUO20} by achieving ~21\% less coefficient of variation (CoV). 

\begin{table*}[h]
\small
\begin{center}
\begin{tabular}{| c | c c | c c | c c | c c | c c |}
\hline 
Performance& 4 days & 4 days & 6 days& 6 days & 9 days & 9 days&  15 days & 15days & 30 days & 30 days \\ 
Metric &  PECAD  & CGNN & PECAD & CGNN & PECAD  & CGNN & PECAD & CGNN & PECAD & CGNN  \\ \hline \hline 
RMSE & - & 231.62 & - & 269.14 & - & 308.12 & - & 460.65 & - & 532.29  \\ \hline
MAE & - & 142.92 & - & 171.33 & - & 207.08 & - & 347.75 & - & 411.25  \\ \hline
CoV & 21.62 & {\bf 16.8} & 24.20 & {\bf 19.5} & 28.46 & {\bf 22.3} & - & 33.4 & - & 38.3 \\ \hline 
R2 value & - & 0.924 & - & 0.897 & - & 0.863 & - & 0.689 & - & 0.569 \\ \hline 
Pearson's & - & 0.962 & - & 0.948 & - & 0.937 & - & 0.885 & - & 0.838 \\ \hline 
\end{tabular}
\end{center}
\caption{Performance metrics for PECAD \cite{GUO20}  and CGNN (our approach) for tomato crop price prediction ({\bf -} means results not available).}
\label{table:TomatoAY}
\end{table*}

\subsection{Results for Potato}
Table \ref{table:PotatoGNN} shows the results obtained for potato with two approaches - PECAD \cite{GUO20} and CGNN (our approach). The paper \cite{GUO20} does not report any results for potato, so we ran the PECAD code (available from the authors of \cite{GUO20}) on our potato dataset. We computed five different performance measures root mean square error, mean absolute error, coefficient of variation, R2 value, and Pearson's correlation coefficient. The results in Table \ref{table:PotatoGNN} clearly show that the CGNN approach outperforms the PECAD approach on all performance metrics for all time horizons. 
Our models are achieving an average of ~27\% less CoV than the PECAD model.

\begin{table*}[h]
\small
\begin{center}
\begin{tabular}{| c | c c | c c | c c | c c | c c |}
\hline 
Performance& 4 days & 4 days & 6 days& 6 days & 9 days & 9 days&  15 days & 15days & 30 days & 30 days \\ 
Metric &  {PECAD}  & {CGNN} & {PECAD} & {CGNN} & PECAD  & CGNN & PECAD & CGNN & PECAD & CGNN \\ \hline \hline 
RMSE & 134.82 & 119.58 & 174.18 & 131.37 & 182.29 & 141.89 & 228.83 & 149.88 & 343.50 & 201.77 \\ \hline
MAE & 92.07 & 66.69 & 138.13 & 77.78 & 142.12 & 89.49 &  184.73 & 100.66 & 269.09 & 148.87 \\ \hline
CoV & 11.1 & 9.8 & 14.4 & 10.7 & 15 & 11.5 & 18.6 & 12.0 & 27.9 & 15.9 \\ \hline 
R2 value & 0.958 & 0.967 & 0.929 & 0.960 & 0.922 & 0.952 & 0.875 & 0.946 & 0.714 & 0.898 \\ \hline 
Pearson's & 0.983 & 0.983 & 0.979 & 0.980 & 0.973 & 0.976 & 0.961 & 0.973 & 0.893 & 0.952 \\ \hline 
\end{tabular}
\end{center}
\caption{Performance metrics for PECAD \cite{GUO20} and CGNN (our approach)  for potato crop price prediction}
\label{table:PotatoGNN}
\end{table*}

\section{Conclusion and Future Work}
The key contribution of this work is to come up with an innovative deep learning model, combining convolutional neural networks and graph neural networks,  to obtain highly accurate estimates of crop prices over different time horizons. We carried out our experimentation on tomato and potato crops across all the markets in India. Our approach produces a performance that is at least 20\% better than the results available in the literature.  One immediate direction for future work is to further optimize the CNN-GNN model and explore other models. Using graph attention networks holds promise in this context.  Another direction is to explore predictions over longer time horizons such as two to three months. We are also currently looking into the formal explainability of the model used in this paper.

\section*{Acknowledgments}
The first author would like to thank the Government of India, Ministry of Education, for providing the doctoral fellowship. All the authors would like to thank the National Bank for Agriculture and Rural Development (NABARD), government of India, for supporting this work.

\bibliographystyle{unsrt}  
\bibliography{AGRIBIB} 

\end{document}